\documentclass[12pt,letterpaper]{article}
\usepackage{authblk} % authors and affiliation
\usepackage[comma,authoryear]{natbib} % nice referencing
\usepackage{amssymb}
\usepackage{latexsym}
\usepackage{graphicx}
\usepackage{url}
\usepackage{array}
\usepackage{color,soul} % highlighting
\usepackage{float} % stop figure from floating with [H] (package {here} is obsolete
\usepackage[colorinlistoftodos]{todonotes}
\usepackage{fancyhdr} % running titles
\usepackage{amsmath}
\usepackage{amssymb}
\usepackage{amsmath}
\usepackage{graphicx}
\usepackage{url}
\usepackage{array}
\usepackage{subfig}
\usepackage{color,soul} % highlighting
\soulregister\ref{7} % make soul work with references
\soulregister\cite{7} % make soul work with citations
\usepackage{float} % stop figure from floating with [H] (package {here} is obsolete
\usepackage{hyperref}
\usepackage{booktabs} % To thicken table lines
\usepackage{rotating}
\usepackage{multirow}
\usepackage{transparent}
\usepackage[subtle,margins=normal,leading=normal]{savetrees} % conservativly saving space (for MICCAI)
\usepackage{bm} % bold math symbols
\usepackage{colortbl}
\usepackage{tabulary}

\newcommand{\argmax}{\operatornamewithlimits{\textbf{arg\,max}}}

\usepackage{geometry}
\usepackage{layout}
\begin{document} %\layout
\clearpage
\newpage % removes top white space
\title{Hierarchical 3D fully convolutional networks for multi-organ segmentation}
\author[1]{\small Holger R. Roth}
\author[1]{\small Hirohisa Oda}
\author[1]{\small Yuichiro  Hayashi}
\author[1]{\small Masahiro Oda}
\author[1]{\small Natsuki  Shimizu}
\author[2]{\small Michitaka Fujiwara}
\author[3]{\small Kazunari Misawa}
\author[1]{\small Kensaku Mori}
\affil[1]{\small Nagoya University, Japan}
\affil[2]{\small Nagoya University Graduate School of Medicine, Japan}
\affil[3]{\small Aichi Cancer Center, Japan}
\date{}
%%%%%%%%%%%%%%%%%%%%%%%%%%%%%%%%%%%%%%%%%%%%%%%%%%%%%%%%%%%%%%%%%%%%%%%%%%%%%%%%%%%%%%%
\maketitle
%%%%%%%%%%%%%%%%%%%%%%%%%%%%%%%%%%%%%%%%%%%%%%%%%%%%%%%%%%%%%%%%%%%%%%%%%%%%%%%%%%%%%%%
\begin{abstract} 
\noindent Recent advances in 3D fully convolutional networks (FCN) have made it feasible to produce dense voxel-wise predictions of full volumetric images. In this work, we show that a multi-class 3D FCN trained on manually labeled CT scans of seven abdominal structures (artery, vein, liver, spleen, stomach, gallbladder, and pancreas) can achieve competitive segmentation results, while avoiding the need for handcrafting features or training organ-specific models. To this end, we propose a two-stage, coarse-to-fine approach that trains an FCN model to roughly delineate the organs of interest in the first stage (seeing $\sim$40\% of the voxels within a simple, automatically generated binary mask of the patient's body). We then use these predictions of the first-stage FCN to define a candidate region that will be used to train a second FCN. This step reduces the number of voxels the FCN has to classify to $\sim$10\% while maintaining a recall high of $>$99\%. This second-stage FCN can now focus on more detailed segmentation of the organs. We respectively utilize training and validation sets consisting of 281 and 50 clinical CT images. Our hierarchical approach provides an improved Dice score of 7.5 percentage points per organ on average in our validation set. We furthermore test our models on a completely unseen data collection acquired at a different hospital that includes 150 CT scans with three anatomical labels (liver, spleen, and pancreas). In such challenging organs as the pancreas, our hierarchical approach improves the mean Dice score from 68.5 to 82.2\%, achieving the highest reported average score on this dataset.
\end{abstract}
%%%%%%%%%%%%%%%%%%%%%%%%%%%%%%%%%%%%%%%%%%%%%%%%%%%%%%%%%%%%%
%%%%%%%%%%%%%%%%%%%%%%%%%%%%%%%%%%%%%%%%%%%%%%%%%%%%%%%%%%%%%
\section{Introduction}
Recent advantages in fully convolutional networks (FCN) have made it feasible to train models for pixel-wise segmentation in an end-to-end fashion \cite{long2015fully}. Efficient implementations of 3D convolution and growing GPU memory have made it feasible to extent these methods to 3D medical imaging and train networks on large amounts of annotated volumes. One such example is the recently proposed 3D U-Net \cite{cciccek20163d}, which applies a 3D FCN with skip connections to sparsely annotated biomedical images. Alternative architectures such as V-Net \cite{milletari2016v} or VoxResNet \cite{chen2016voxresnet} have also been successfully applied to 3D medical images.
%%%%%%%%%%%%%%%%%%%%%%%
In this work, we show that a 3D FCN trained on manually labeled data of seven abdominal structures can also achieve competitive segmentation results on clinical CT images. Our approach applies the 3D FCN architectures to problems of multi-organ segmentation in a hierarchical fashion. An FCN can be trained on whole 3D CT scans, however, because of the high imbalance between background and foreground voxels (organs, vessels, etc.), many neurons will only learn to differentiate foreground from the background voxels in order to minimize the loss function used for training the network. While this enables the FCN to roughly segment the organs, it causes particularly smaller organs (like the pancreas or gallbladder) and vessels to suffer from inaccuracies around their boundaries \cite{roth2017spatial,zhou2016pancreas}. 

\begin{figure}[htb]
  \centering
   \includegraphics[width=0.75\linewidth, clip]{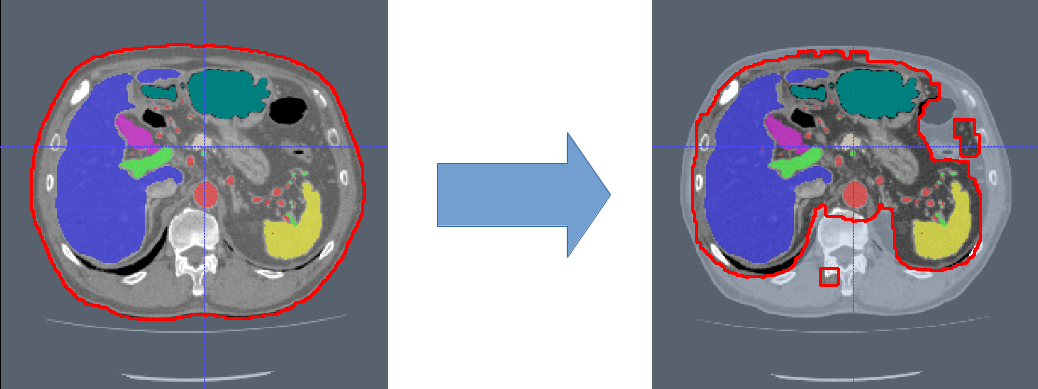}
   \caption{\small Hierarchical 3D fully convolutional networks in a coarse-to-fine approach: stage 1 (left) learns the generation of a candidate region for training second-stage FCN (right) for finer prediction.  Outlined red area shows candidate region $C_1$ used in first stage and $C_2$ used in second stage. Colored regions denote ground truth annotations for training (best viewed in color).}
   \label{fig:coarse-to-fine}
\end{figure}

%%%%%%%%%%%%%%%%%%%%%%%
To overcome this limitation, we learn a second-stage FCN that is fine-tuned from a first-stage FCN in a hierarchical manner and hence focuses more on the boundary regions. This is a coarse-to-fine approach in which the first-stage FCN sees around 40\% of the voxels using only a simple mask of the body created by thresholding the image. In the second stage, the amount of the image\rq{}s voxels is reduced by around 10\%. In effect, this step narrows down and simplifies the search space for the FCN to decide which voxels belong to the background or any of the foreground classes; this strategy has been successful in many computer vision problems. Our approach is illustrated on a training example in Fig. \ref{fig:coarse-to-fine}. 
%%%%%%%%%%%%%%%%%%%%%%%
In contrast to previous approaches of multi-organ segmentation where separate models have to be created for each organ \cite{oda2016regression,tong2015discriminative}, our proposed method allows us to use the same model to segment such very different anatomical structures as large abdominal organs (liver, spleen), but also vessels like arteries and portal veins. In contrast, other recent FCN-based methods are applied to medical imaging were often constrained to using rectangular bounding boxes around single organs and performing slice-wise processing in 2D \cite{roth2017spatial,zhou2016pancreas,christ2016automatic}.
%%%%%%%%%%%%%%%%%%%%%%%%%%%%%%%%%%%%%%%%%%%%%%%%%%%%%%%%%%%%%
%%%%%%%%%%%%%%%%%%%%%%%%%%%%%%%%%%%%%%%%%%%%%%%%%%%%%%%%%%%%%
\section{Methods}
Convolutional neural networks have the ability to solve challenging classification tasks in a data-driven manner. Given a training set of images and labels $\mathbf{S} = \left\{(I_n,L_n), n = 1,\dots,N\right\}$, $I_n$ denotes the raw CT images and $L_n$ denotes the ground truth label images. In our case, each $L_n$ contains $K=8$ class labels consisting of the manual segmentations of seven organs (artery, portal vein, liver, spleen, stomach, gallbladder, and pancreas) and the background for each voxel in the CT image. The employed network architecture is the 3D extension by {\c{C}}i{\c{c}}ek et al. \cite{cciccek20163d} of the U-Net proposed by Ronneberger et al. \cite{ronneberger2015u}. U-Net, which is a type of fully convolutional network (FCN) \cite{long2015fully} optimized for bio-medical image applications, utilizes up-convolutions to remap the lower resolution feature maps within the network to the denser space of the input images. This operation allows for denser voxel-to-voxel predictions in contrast to previously proposed sliding-window CNN methods where each voxel under the window is classified independently making such architecture inefficient for processing large 3D volumes. In 3D U-Nets, such operations as 2D convolution, 2D max-pooling, and 2D up-convolution are replaced by their 3D counterparts \cite{cciccek20163d}. We use the open-source implementation of 3D U-Net\footnote{\url{http://lmb.informatik.uni-freiburg.de/resources/opensource/unet.en.html}} based on the Caffe deep learning library \cite{jia2014caffe}. 3D U-Net architecture consists of analysis and synthesis paths with four resolution levels each. Each resolution level in the analysis path contains two $3 \times 3 \times 3$ convolutional layers, each followed by rectified linear units (ReLu) and a $2 \times 2 \times 2$ max pooling with strides of two in each dimension. In the synthesis path, the convolutional layers are replaced by up-convolutions of $2 \times 2 \times 2$ with strides of two in each dimension. These are followed by two $3 \times 3 \times 3$ convolutions, each of which has a ReLu. Furthermore, 3D U-Net employs shortcut (or skip) connections from layers of equal resolution in the analysis path to provide higher-resolution features to the synthesis path \cite{cciccek20163d}. The last layer contains a $1\times 1\times 1$ convolution that reduces the number of output channels to the number of class labels (which is $K=8$ in our case) and a size of $44 \times 44 \times 28$ of each channel. This architecture has over 19 million parameters \cite{cciccek20163d}. The model can be trained to minimize weighted voxel-wise cross-entropy loss: 
\begin{equation} \small
	\mathcal{L} \ = \ \frac{-1}{N}\sum^{K}_{i=1} \lambda_i \times \left( \sum_{x\forall N_i}\log{\left(\hat{p}_{k}(x)\right)} \right),
	\label{equ:loss}
\end{equation}
where $\hat{p}$ are the softmax output class probabilities
\begin{equation} \small
	\hat{p}_{k}(x) \ = \ \frac{\exp(x_{k}(x))}{    \sum^{K-1}_{k\rq{}=1}\exp(x_{k\rq{}}(x))       },
\end{equation} 
and $\lambda_i$ is a weight factor (Eq. \ref{equ:weight}), $N$ are the total number of voxels $x$, $N_i$ are the number of voxels within one class in $L_n$, and $k \in [0,1,2,...,K-1]$ indicates the correct ground truth label. The input to this loss function is real valued predictions $x \in [-\infty,+\infty]$ from the last convolutional layer. We apply a voxel-wise weight $\lambda_i$ to the loss function (Eq. \ref{equ:loss}) to balance the common voxels (i.e., background) with respect to such smaller organs as vessels or the pancreas. We choose $\lambda_i$ such that $\sum^{K}_{i=1} \lambda_i = 1$, with
\begin{equation} \small
	\lambda_i = \frac{1-N_i/N_C}{K-1},
	\label{equ:weight}
\end{equation}
where $N_C$ is the number of voxels within candidate region $C_1$ or $C_2$.
%%%%%%%%%%%%%%%%%%%%%%%%%%%%%%%%%%%%%%%%%%%%%%%%%%
\subsection{Coarse-to-fine prediction}
%%%%%%%%%%%%%%%%%%%%%%%%%%%%%%%%%%%%%%%%%%%%%%%%%%
Due to GPU memory restrictions, the network input is fixed to $132 \times 132 \times 116$ sub-volumes that are randomly sampled from the candidate regions within the training CT images, as described below. To increase the field of view presented to the CNN and reduce the redundancy among neighboring voxels, each image is downsampled by a factor of 2. The resulting prediction maps are then resampled back to the original resolution using nearest neighbor interpolation. 

\begin{figure}[htb]
  \centering
  \subfloat[Training]{\includegraphics[width=0.45\textwidth]{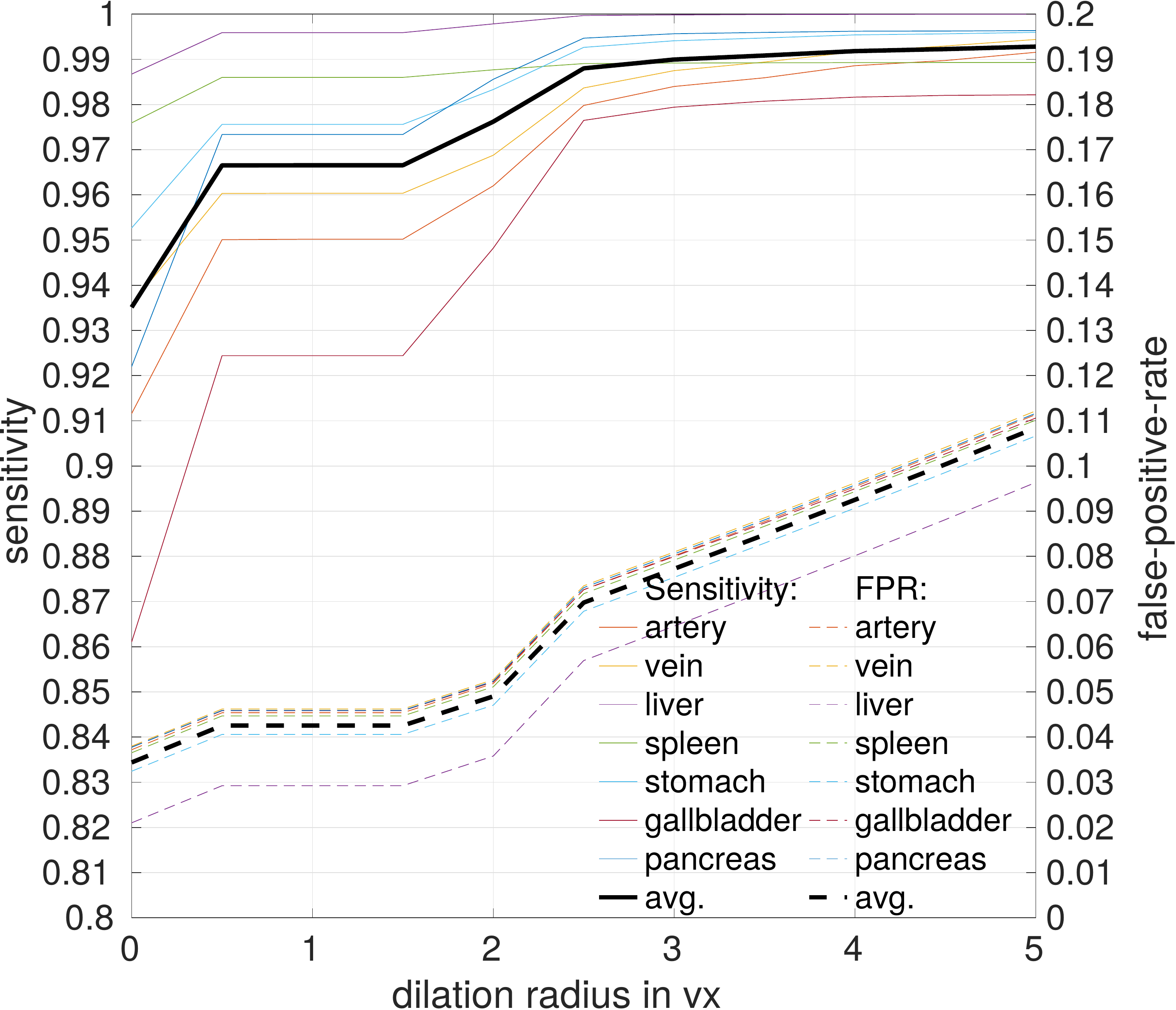}}
    %\ \ \ \
   \hfill
  \subfloat[Validation]{\includegraphics[width=0.45\textwidth]{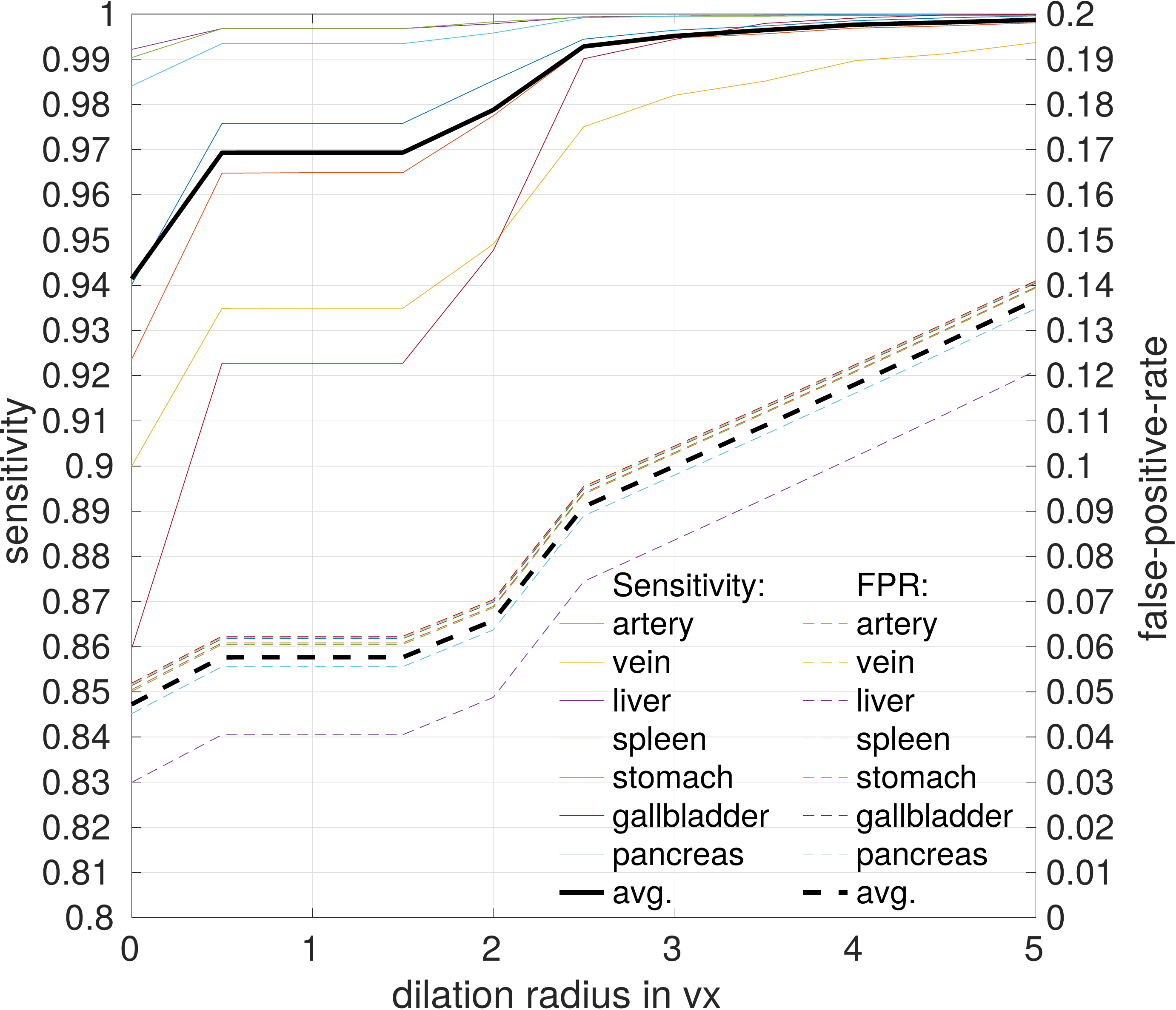}}    
   \caption{\small Sensitivity and false-positive-rate (FPR) as a function of dilating prediction maps of first stage in training (a) and validation (b). We observe good trade-off between high sensitivity ($>$99\% on average) and low false-positive-rate ($\sim$10\% on average) at dilation radius of $r=3$.}
   \label{fig:recall_stage1}
\end{figure}

In the first stage, we apply simple thresholding in combination with morphological operations (hole filling and largest component selection) to get a mask of the patient's body in a slice-by-slice fashion. This mask can be utilized as candidate region $C_1$ to reduce the number of voxels necessary to compute the network's loss function and reduce the amount of input 3D regions shown to the CNN during training to about 40\%. After training, the first-stage FCN is applied to each image to generate candidate regions $C_2$ for training the second-stage FCN (Fig. \ref{fig:prediction_rendering}). We define the organ labels in the testing phase using the $\argmax$ of the class probability maps. Any foreground label is then dilated in 3D using a voxel radius of $r$. We compare the recall and false-positive rates of this first-stage FCN with respect to $r$ for both the training and validation sets in Fig. \ref{fig:recall_stage1}. $r=3$ gives good trade-off between high recall ($>$99\%) and low false-positive rates for each organ on our training and validation sets.
%%%%%%%%%%%%%%%%%%%%%%%%%%%%%%%%%%%%%%%%%%%%%%%%%%

\textbf{Training}: The network iteratively adjusts its parameters by stochastic gradient descent. Batch normalization is used throughout the network for improved convergence \cite{ioffe2015batch}. Furthermore, we utilize random elastic deformations in 3D during training to artificially increase the amount of available data samples, as proposed by \cite{cciccek20163d}. Each training sub-volume is randomly extracted from $C_1$ or $C_2$. 
%%%%%%%%%%%%%%%%%%%%%%%%%%%%%%%%%%%%%%%%%%%%%%%%%%

\textbf{Testing}: The CT image is processed by the 3D FCN using a tiling strategy (sliding-window) \cite{cciccek20163d}. For greater speed, we use non-overlapping tiles in the first stage and investigate the use of non-overlapping and overlapping tiles in the second. When using overlapping tiles (with a $R=4$ times higher sampling rate of each voxel $x$), the resulting probabilities for the overlapping voxels are averaged:% $p = \frac{1}{R}\sum^{R}_{r=1}\hat{p}_r$.
\begin{equation} \small
     p(x) = \frac{1}{R}\sum^{R}_{r=1} \ p_r(x).
	\label{equ:overlapping}
\end{equation}
%%%%%%%%%%%%%%%%%%%%%%%%%%%%%%%%%%%%%%%%%%%%%%%%%%%%%%%%%%%%%
%%%%%%%%%%%%%%%%%%%%%%%%%%%%%%%%%%%%%%%%%%%%%%%%%%%%%%%%%%%%%
%%%%%%%%%%%%%%%%%%%%%%%%%%%%%%%%%%%%%%%%%%%%%%%%%%%%%%%%%%%%%
\section{Experiments \& Results}
%%%%%%%%%%%%%%%%%%%%%%%%%%%%%%%%%%%%%%%
\begin{figure}[htb] %0.32,0.32,0.32
  \centering
  \subfloat[Ground truth]{\includegraphics[width=0.3\textwidth]{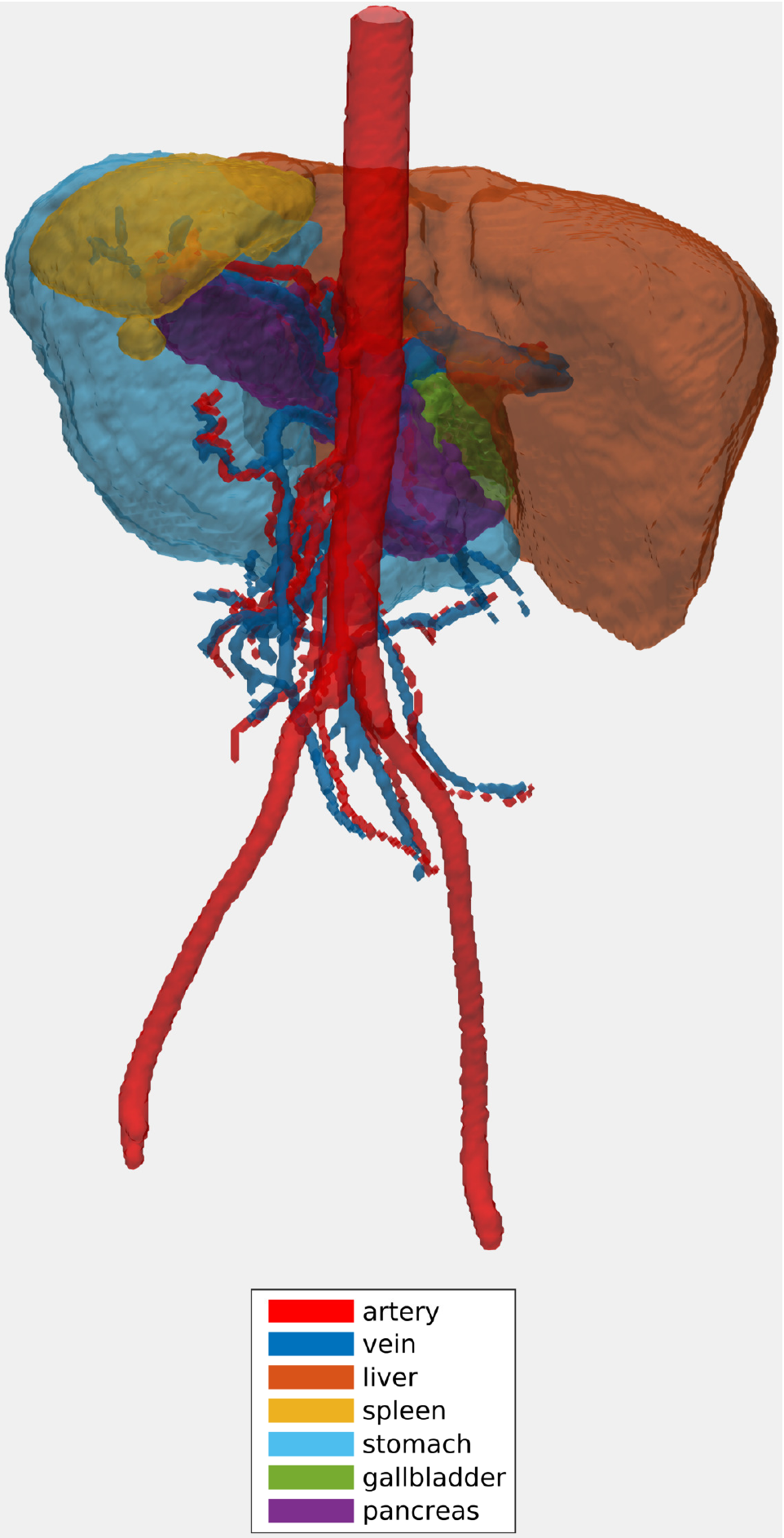}}
  \hfill
  \subfloat[Stage 2 - Tiling]{\includegraphics[width=0.3\textwidth]{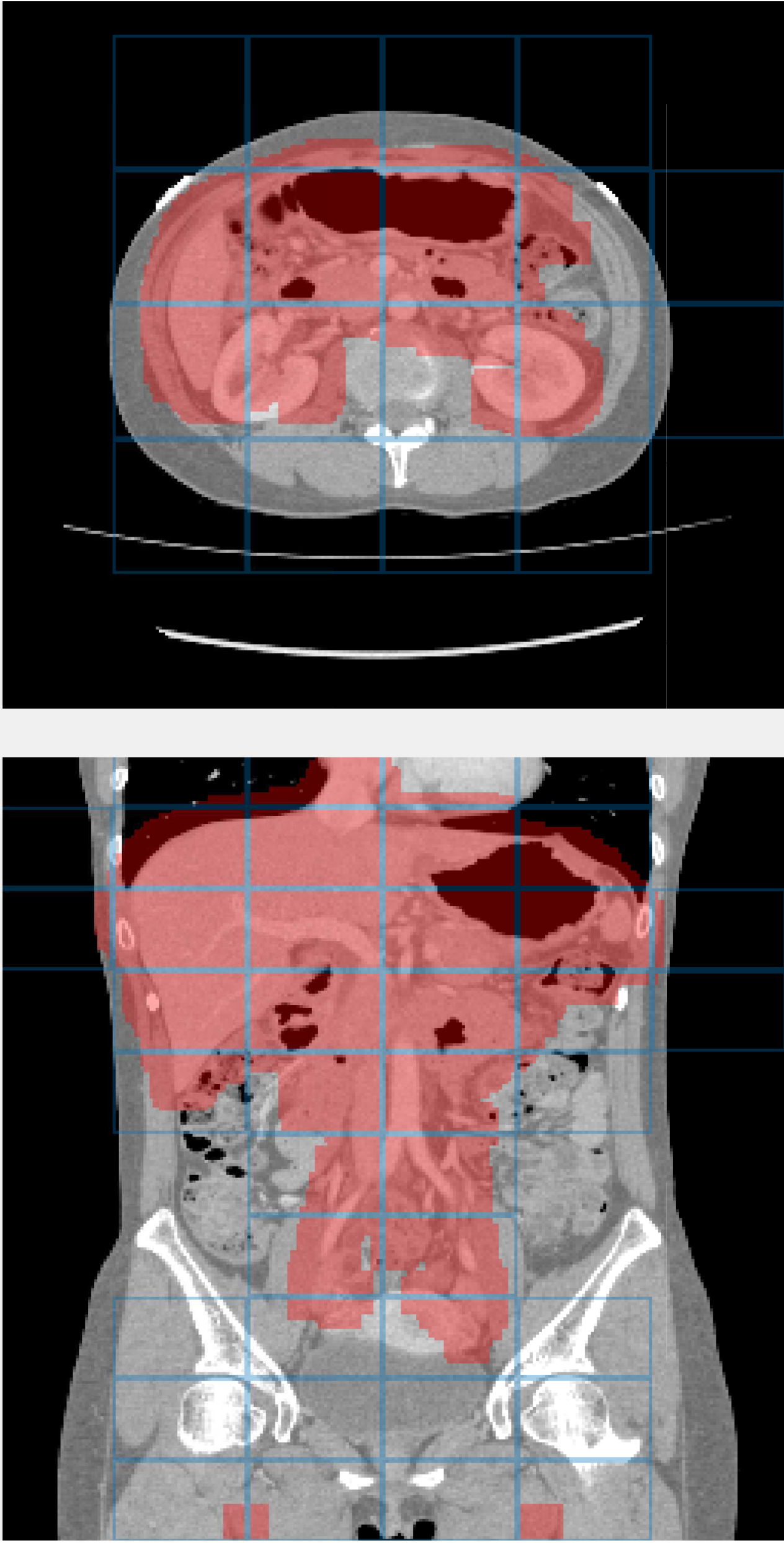}}    
  \hfill
  \subfloat[Stage 2 - N/OL]{\includegraphics[width=0.3\textwidth]{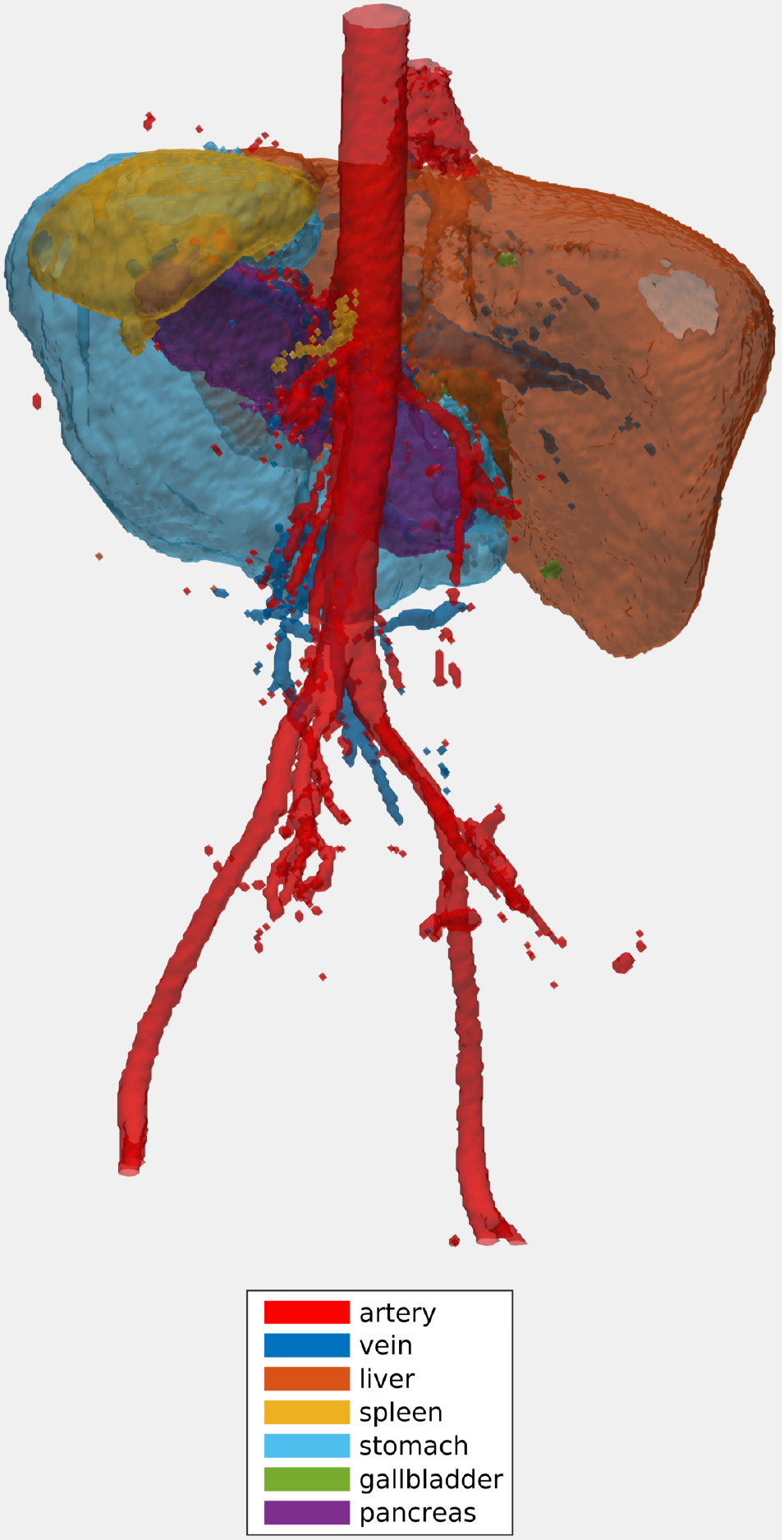}}    
   \caption{\small Example of the validation set with (a) ground truth and illustrating (b) non-overlapping (N/OL) tiling approach on 2nd stage candidate region $C_2$. Resulting segmentation is shown in (c). Note that the grid shows the output tiles of size $44 \times 44 \times 28$ ($x,y,z$-directions). Each predicted tile is based on a larger input of $132 \times 132 \times 116$ that the network processes.}
   \label{fig:prediction_rendering}
\end{figure}
\begin{figure}[htb] %0.235, 0.24, 0.24, 0.24
  \centering
  \subfloat[Ground truth]{\includegraphics[width=0.238\textwidth]{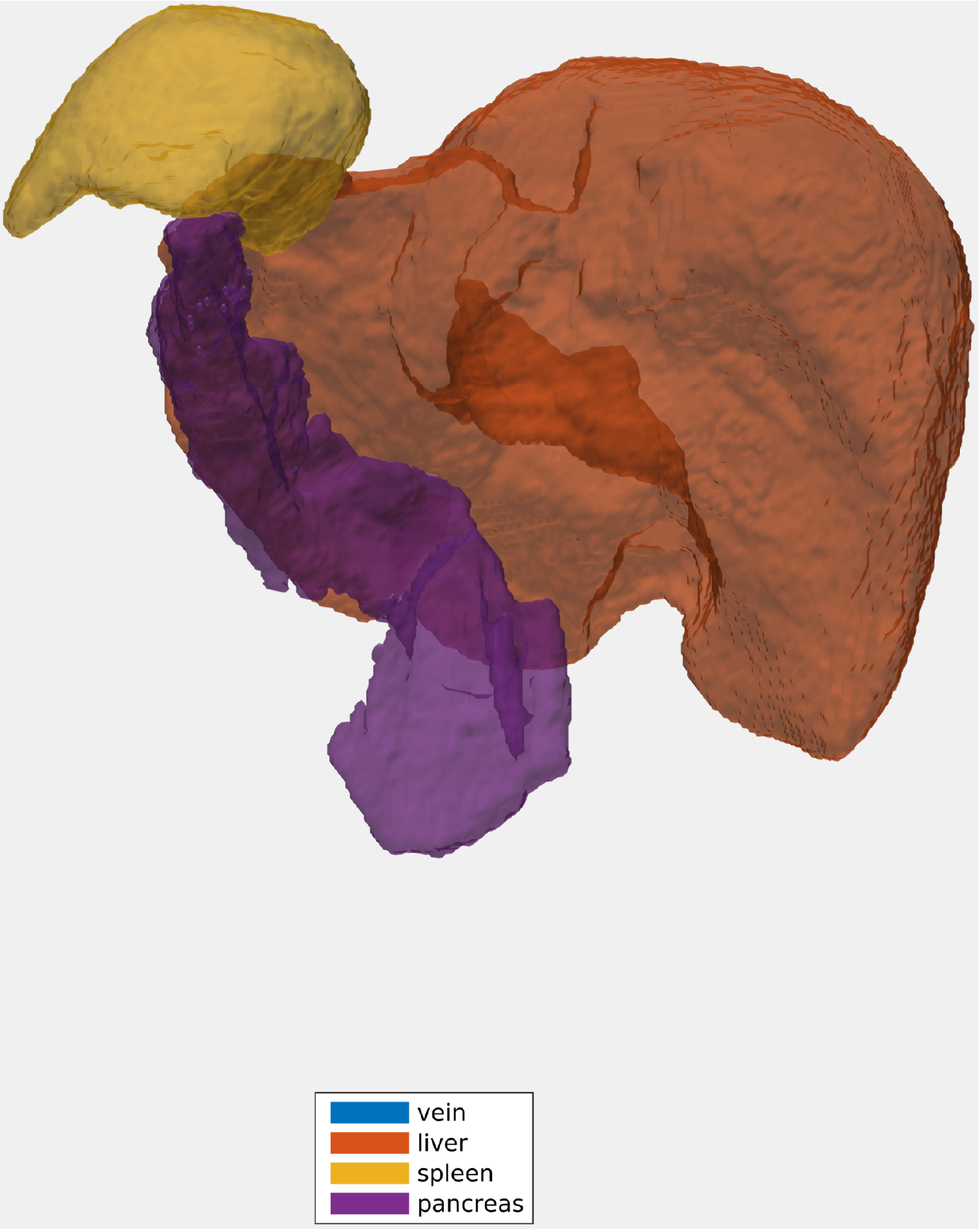}}
  \hfill
  \subfloat[Stage 1]{\includegraphics[width=0.245\textwidth]{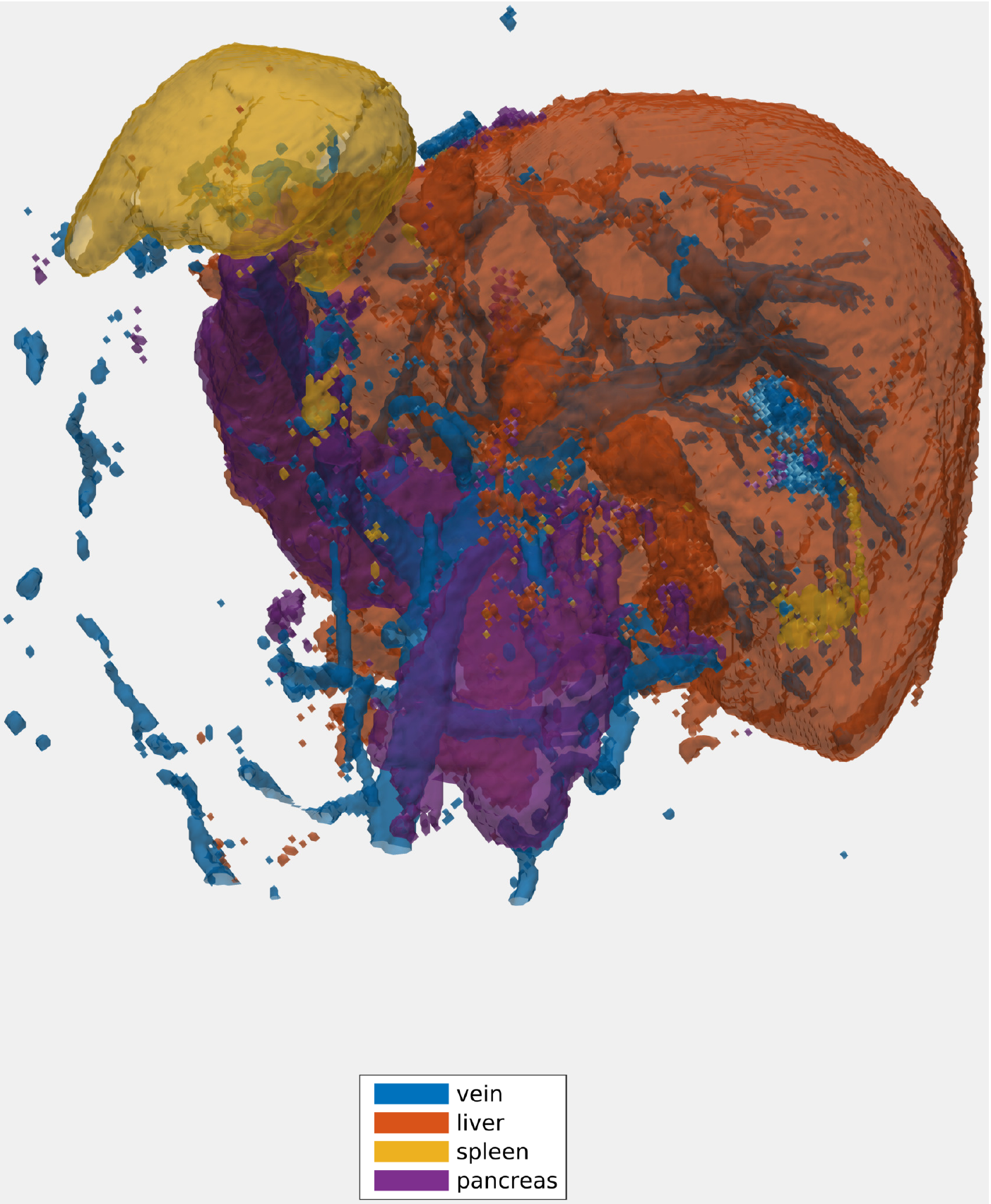}}  
  \hfill
  \subfloat[Stage 2: N/OL]{\includegraphics[width=0.245\textwidth]{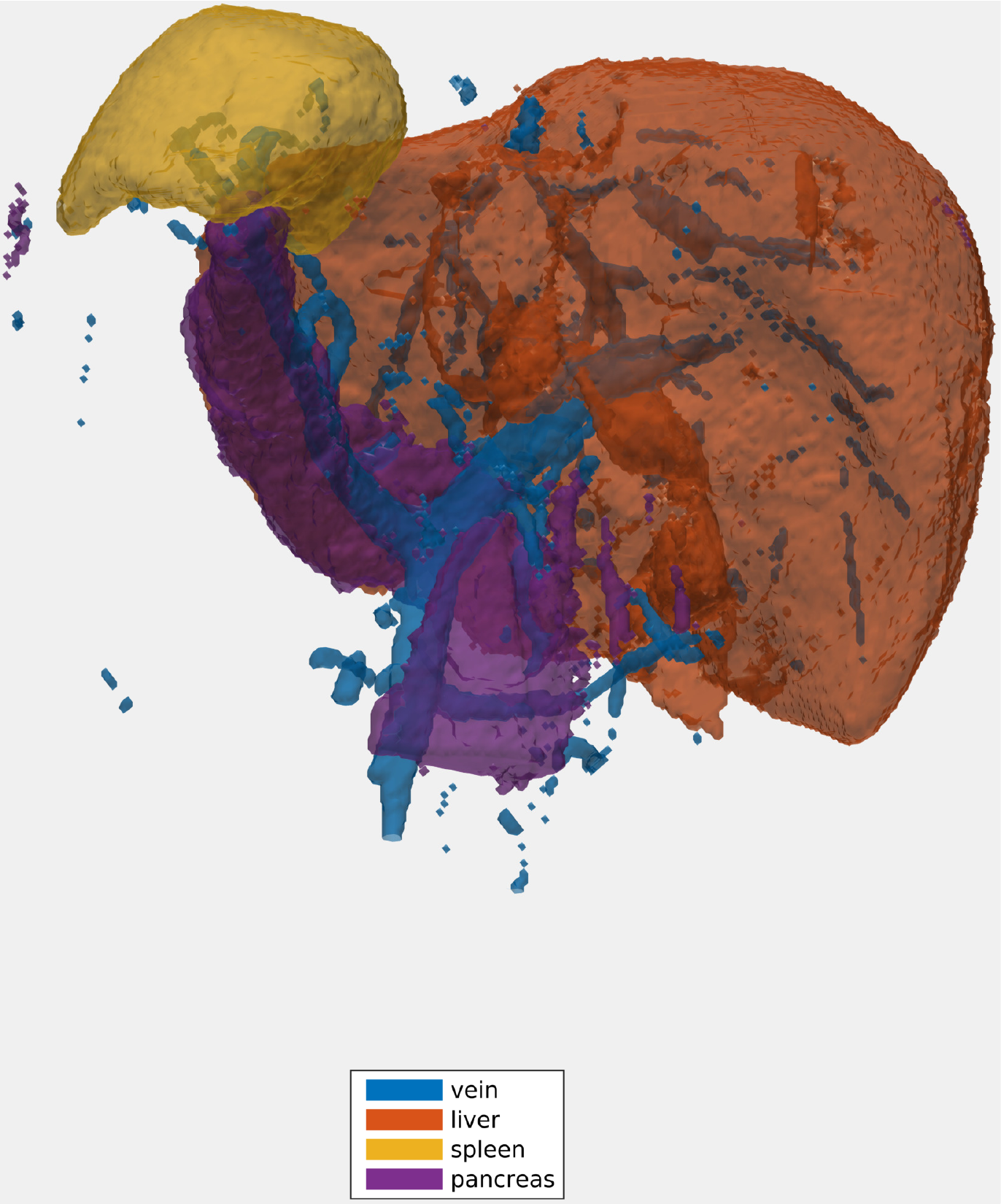}}    
  \hfill
  \subfloat[Stage 2: OL]{\includegraphics[width=0.245\textwidth]{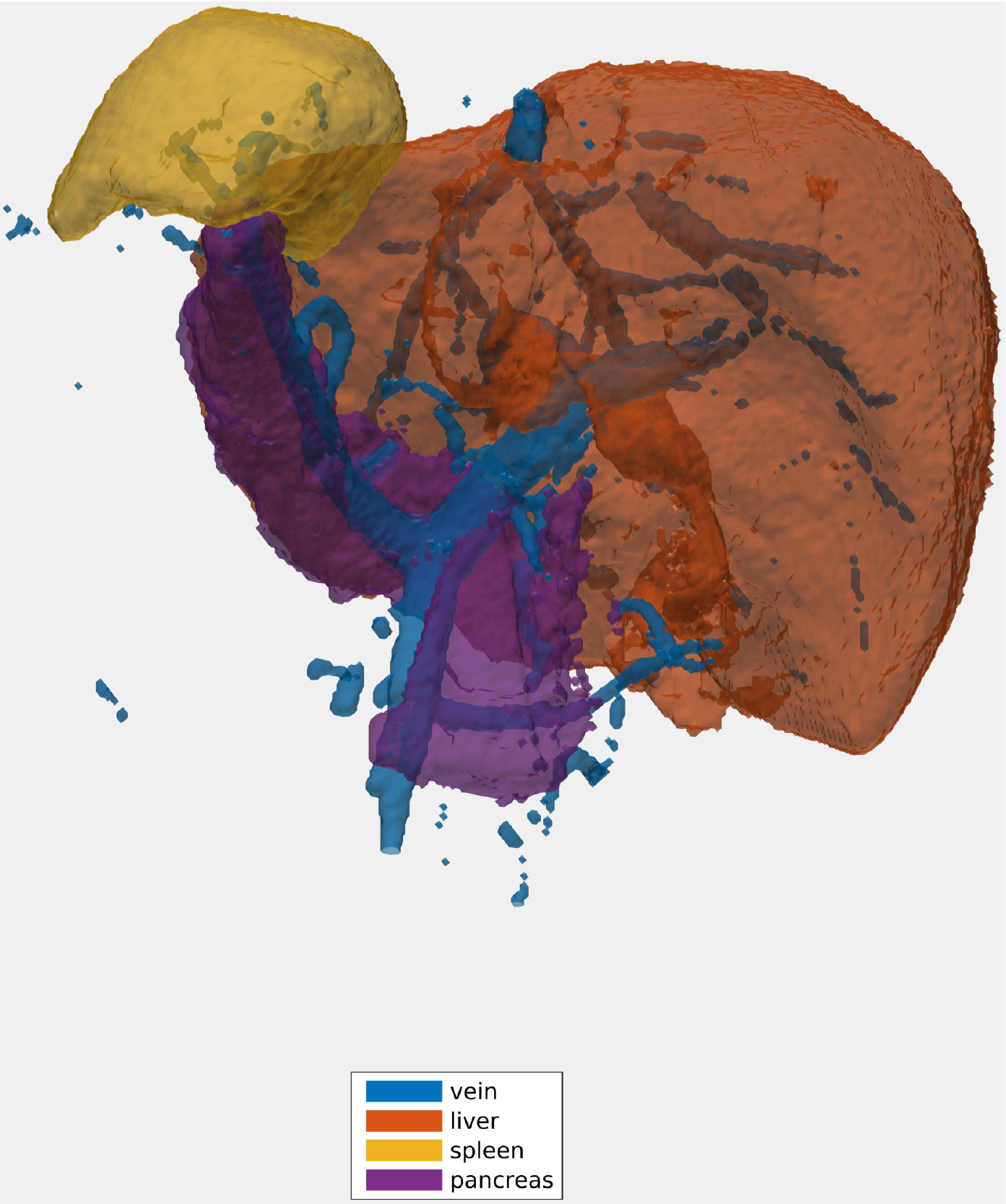}}        
   \caption{\small Surface renderings: 
   (a) ground truth segmentation, 
   (b) result of proposed method in first stage, second-stage results using (c) non-overlapping (N/OL), and (d) overlapping (OL) tiles strategy.}
   \label{fig:surface_stages}
\end{figure}

%%%%%%%%%%%%%%%%%%%%%%%%%%%%%%%%%%%%%%%
\textbf{Training and validation:} Our dataset includes 331 contrast-enhanced abdominal clinical CT images in the portal venous phase used for pre-operative planning in gastric surgery. Each CT volume consists of $460-1177$ slices of $512\times 512$ pixels. The voxel dimensions are [0.59-0.98, 0.59-0.98, 0.5-1.0] mm. A random split of 281/50 patients is used for training and validating the network, i.e., determining when to stop training to avoid overfitting. In both training stages, we employ smooth B-spline deformations to both the image and label data, as proposed by \cite{cciccek20163d}. The deformation fields are randomly sampled from a normal distribution with a standard derivation of 4 and a grid spacing of 32 voxels. Furthermore, we applied random rotations between $-5^\circ$ and $+5^\circ$ to the training images for plausible deformations during training. No deformations were applied during the testing. We trained 200,000 iterations in the first stage and 115,000 in the second. Table \ref{tab:dice_results_validation} summarizes the Dice similarity scores for each organ labeled in the 50 validation cases. On average, we achieved a 7.5\% improvement in Dice scores per organ. Small, thin organs such as arteries especially benefit from our two-stage hierarchical approach. For example, the mean Dice score for arteries improved from 59.0 to 79.6\% and from 54.8 to 63.1\% for the pancreas. The effect is less pronounced for large organs, like the liver, the spleen, and the stomach. Fig. \ref{fig:prediction_rendering} shows an example result from the validation set and illustrates the tiling approach. The 3D U-Net separates the foreground organs well from the background tissue of the images. 
%%%%%%%%%%%%%%%%%%%%%%%%%%%%%%
%%%%% ACC %%%%%%%%%%%%%%%%%%%%%%
%%%%% %%%%%%%%%%%%%%%%%%%%%%%%%
\begin{table}
\scriptsize
\centering
\caption{\small \textbf{Validation set:} Dice similarity score [\%] of different stages of FCN processing}
\label{tab:dice_results_validation}
\begin{tabular}{l|rrrrrrr|r}
\multicolumn{8}{l}{\textbf{Stage 1: Non-overlapping}} & \tabularnewline
\hline
\rowcolor[gray]{.9}\textbf{Dice} & \textbf{artery} & \textbf{\ \ \ \ vein} & \textbf{\ \ \ liver} & \textbf{\ \ spleen} & \textbf{\ stomach} & \textbf{\ gallbladder} & \textbf{\ pancreas} & \textbf{ \ \ Avg}\tabularnewline
\hline
\textbf{Mean} & 59.0 & 64.7 & 89.6 & 84.1 & 80.0 & 69.6 & 54.8 & 71.7\tabularnewline
\rowcolor[gray]{.9}\textbf{Std} & 7.8 & 8.6 & 1.7 & 4.7 & 18.3 & 14.1 & 11.0 & 9.5\tabularnewline
\textbf{Median} & 59.8 & 67.3 & 90.0 & 85.2 & 87.5 & 73.2 & 57.2 & 74.3\tabularnewline
\rowcolor[gray]{.9}\textbf{Min} & 41.0 & 34.5 & 84.4 & 70.9 & 8.4 & 13.8 & 23.5 & 39.5\tabularnewline
\textbf{Max} & 75.7 & 76.0 & 92.6 & 91.4 & 94.8 & 86.8 & 72.0 & 84.2\tabularnewline
\hline
 &  &  &  &  &  &  &  & \tabularnewline
\multicolumn{8}{l}{\textbf{Stage 2: Non-overlapping}} & \tabularnewline
\hline
\rowcolor[gray]{.9}\textbf{Dice} & \textbf{artery} & \textbf{vein} & \textbf{liver} & \textbf{spleen} & \textbf{stomach} & \textbf{gallbladder} & \textbf{pancreas} & \textbf{Avg}\tabularnewline
\hline
\textbf{Mean} & 79.6 & 73.1 & 93.2 & 90.6 & 84.3 & 70.6 & 63.1 & 79.2\tabularnewline
\rowcolor[gray]{.9}\textbf{Std} & 6.5 & 7.9 & 1.5 & 2.8 & 17.3 & 15.9 & 10.7 & 8.9\tabularnewline
\textbf{Median} & 82.3 & 74.6 & 93.5 & 91.2 & 90.9 & 77.3 & 64.5 & 82.1\tabularnewline
\rowcolor[gray]{.9}\textbf{Min} & 62.9 & 33.3 & 88.9 & 82.3 & 10.9 & 13.0 & 32.4 & 46.2\tabularnewline
\textbf{Max} & 87.0 & 83.2 & 95.6 & 95.1 & 96.3 & 89.4 & 81.8 & 89.8\tabularnewline
\hline
% &  &  &  &  &  &  &  & \tabularnewline
%\multicolumn{8}{l}{\textbf{Stage2 vs Stage1}} & \tabularnewline
%\hline
%\rowcolor[gray]{.9}\textbf{Dice} & \textbf{artery} & \textbf{vein} & \textbf{liver} & \textbf{spleen} & \textbf{stomach} & \textbf{gallbladder} & \textbf{pancreas} & %\textbf{Avg}\tabularnewline
%\hline
%\textbf{Mean} & 20.61 & 8.41 & 3.60 & 6.42 & 4.22 & 0.93 & 8.26 & 7.49\tabularnewline
%\rowcolor[gray]{.9}\textbf{Std} & -1.24 & -0.68 & -0.18 & -1.97 & -0.97 & 1.78 & -0.35 & -0.52\tabularnewline
%\textbf{Median} & 22.57 & 7.34 & 3.42 & 6.00 & 3.44 & 4.15 & 7.31 & 7.75\tabularnewline
%\rowcolor[gray]{.9}\textbf{Min} & 21.83 & -1.20 & 4.47 & 11.35 & 2.44 & -0.75 & 8.85 & 6.71\tabularnewline
%\textbf{Max} & 11.28 & 7.21 & 3.06 & 3.70 & 1.52 & 2.67 & 9.74 & 5.60\tabularnewline
\end{tabular}
\end{table}
%%%%%%%%%%%%%%%%%%%%%%%%%%%%%%%%%%%%%%%
%%%%%%%%%%%%%%%%%%%%%%%%%%%%%%%%%%%%%%%
\\\indent\textbf{Testing:} Our test set is different from our training and validation data. It originates from a different hospital, scanner, and research study with gastric cancer patients. 150 abdominal CT scans were acquired in the portal venous phase. Each CT volume consists of $263-1061$ slices of $512\times 512$ pixels. Voxel dimensions are [0.55-0.82, 0.55-0.82, 0.4-0.80] mm. The pancreas, liver, and spleen were semi-automatically delineated by three trained researchers and confirmed by a clinician. Figure \ref{fig:surface_stages} shows surface renderings for comparison of the different stages of the algorithm. A typical testing case in the first and second stages is shown using non-overlapping and overlapping tiles (Eq. \ref{equ:overlapping}). Dice similarity scores are listed in Table \ref{tab:dice_results_testing}. %The second stage achieves the highest reported average score on pancreas in this dataset with 82.22\% $\pm$ 10.20\%. Previous state of the art on this dataset was at 75.1\% $\pm$15.4\% \cite{oda2016regression}. 
This dataset provides slightly higher image quality than our training/validation dataset. Furthermore, its field of view is slightly more constrained to the upper abdomen. This probably explains the better performance for liver and pancreas compared to the validation set in Table \ref{tab:dice_results_validation}.
%%%%%%%%%%%%%%%%%%%%%%%%%%%%%%%%
\\\indent\textbf{Computation:} Training on 281 cases can take 2-3 days for 200-k iterations on a NVIDIA GeForce GTX TITAN X with 12 GB memory. However, the execution time during testing is much faster. On 150 cases of the test set, the processing time for each volume was 1.4-3.3 minutes for each stage, depending on the size of the candidate regions. The processing time increased to 1.6-4.4 minutes using overlapping tiles in stage 2. Testing was performed on NVIDIA GeForce 1080s with 8 GB memory.
%%%%%%% NUSURGERY %%%%%%%%%%%%%%%%%%
%%%%%%% %%%%%%%%%%%%%%%%%%%%%%%%%
\begin{table}
%\small
%\footnotesize
\scriptsize
\centering
\caption{\small \textbf{Testing on unseen dataset:} Dice similarity score [\%] of different stages of FCN processing.}
\label{tab:dice_results_testing}
\begin{tabular}{l|rrr|lrrr|lrrr}
& \multicolumn{3}{l}{\textbf{Stage 1: Non-overlapping}} &\ \ \ & \multicolumn{3}{l}{\textbf{Stage 2: non-overlapping}} &\ \ \ & \multicolumn{3}{l}{\textbf{Stage 2: Overlapping}} \tabularnewline
\hline
\rowcolor[gray]{.9}\textbf{Dice} & \textbf{liver} & \textbf{spleen} & \textbf{pancreas} &  & \textbf{liver} & \textbf{spleen} & \textbf{pancreas} &  & \textbf{liver} & \textbf{spleen} & \textbf{pancreas}\tabularnewline
\hline
\textbf{Mean} & 93.6 & 89.7 & 68.5 &  & 94.9 & 91.4 & 81.2 &  & 95.4 & 92.8 & 82.2\tabularnewline
\rowcolor[gray]{.9}\textbf{Std} & 2.5 & 8.2 & 8.2 &  & 2.1 & 8.9 & 10.2 &  & 2.0 & 8.0 & 10.2\tabularnewline
\textbf{Median} & 94.2 & 91.8 & 70.3 &  & 95.4 & 94.2 & 83.1 &  & 96.0 & 95.4 & 84.5\tabularnewline
\rowcolor[gray]{.9}\textbf{Min} & 78.2 & 20.6 & 32.0 &  & 80.4 & 22.3 & 1.9 &  & 80.9 & 21.7 & 1.8\tabularnewline
\textbf{Max} & 96.8 & 95.7 & 82.3 &  & 97.3 & 97.4 & 91.3 &  & 97.7 & 98.1 & 92.2\tabularnewline
\hline
\end{tabular}
\end{table}
%%%%%%%%%%%%%%%%%%%%%%%%%%%%%%%%%%%%%%%
%%%%%%%%%%%%%%%%%%%%%%%%%%%%%%%%
\\\indent\textbf{Comparison to other methods:} Even though direct comparison is difficult due to the differences in datasets, training/testing evaluation schemes, and segmented organs, we try to indicate how well our model performed with respect to recent state-of-the-art methods in Table \ref{tab:comparison}.
%%%%%%%%%%%%%%%%%%%%%%%%%%%%%%%%
% Table generated by Excel2LaTeX from sheet 'Sheet1'
\begin{table}[htbp]
\scriptsize
  \centering
  \caption{\small Comparison to other methods. We list other recent segmentation work performed on the same/similar datasets and organs and based on atlas-based segmentation propagation using 
  %global affine \cite{wang2014geodesic}, 
local non-rigid registration methods \cite{wolz2013automated} and in combination with machine learning (ML) \cite{tong2015discriminative}. We also list a method using regression forest (RF) and graph cut (GC) \cite{oda2016regression}, and two other methods utilizing 2D FCNs \cite{roth2017spatial,zhou2016pancreas}. Validation of other methods was performed using either leave-one-out-validation (LOOV) or cross-validation(CV).}
	\label{tab:comparison}%
    \begin{tabular}{lcllllr}
		\hline
    \rowcolor[gray]{.9}\textbf{Method} & \textbf{Subjects} \ \ \ & \textbf{Approach}  \ \ \ \ \ & \textbf{Validation} \ \ \ \ & \textbf{Organs} \ \ \ \ \ \ & \textbf{Dice [\%]} \ \ \ \ \ & \textbf{Time [h]} \\
		\hline
    Proposed & 150   & 3D FCN & Testing & Liver & 95.4 $\pm$ 2.0 & 0.07 \\
          &       &       &       & Spleen & 92.8 $\pm$ 8.0 &  \\
          &       &       &       & Pancreas & 82.2 $\pm$ 10.2 &  \\
    \rowcolor[gray]{.9}\cite{tong2015discriminative} & 150   & Global affine & LOOV & Liver & 94.9 $\pm$ 1.9 & 0.5 \\
    \rowcolor[gray]{.9}      &       &  + ML &       & Spleen & 92.5 $\pm$ 6.5 &  \\
    \rowcolor[gray]{.9}      &       &       &       & Pancreas & 71.1 $\pm$ 14.7 &  \\
%    Wang et al. \cite{wang2014geodesic} & 100     & Global affine & LOOV & Liver & 94.5 $\pm$ 2.5 & 14 \\
%          &       &       &       & Spleen & 92.5 $\pm$ 8.4 &  \\
%          &       &       &       & Pancreas & 65.5 $\pm$ 18.6 &  \\
    \rowcolor[gray]{.9}\cite{wolz2013automated}  & 150      & Local non-rigid & LOOV & Liver & 94.0 $\pm$ 2.8 & 51 \\
    \rowcolor[gray]{.9}      &       &       &       & Spleen & 92.0 $\pm$ 9.2 &  \\
    \rowcolor[gray]{.9}      &       &       &       & Pancreas & 69.6 $\pm$ 16.7 &  \\
    \cite{oda2016regression} & 147   & RF + GC & LOOV & Pancreas & 75.1 $\pm$ 15.4 & 3 \\
    \rowcolor[gray]{.9}\cite{roth2017spatial} & 82    & 2D FCN & 4-fold CV & Pancreas & 81.3 $\pm$ 6.3 & 0.05 \\
     \cite{zhou2016pancreas} & 82    & 2D FCN & 4-fold CV & Pancreas & 82.4 $\pm$ 5.7 & n/a \\
    \hline
    \end{tabular}%
\end{table}%    
%%%%%%%%%%%%%%%%%%%%%%%%%%%%%%%%%%%%%%%%%%%%%%%%%%%%%%%%%%%%%
%%%%%%%%%%%%%%%%%%%%%%%%%%%%%%%%%%%%%%%%%%%%%%%%%%%%%%%%%%%%%
\section{Discussion \& Conclusion}
The hierarchical coarse-to-fine approach presented in this paper provides a simple yet effective method for employing 3D FCNs in medical imaging settings. No post-processing was applied to any of the FCN output. The improved performance stemming from our hierarchical approach is especially visible in smaller, thinner organs, such as arteries, veins, and pancreas. Note that we used different datasets (from different hospitals and scanners) for separate training/validation and testing. This experiment illustrates our method\rq{}s generalizability and robustness to differences in image quality and populations. Running the algorithms at half resolution efficiently improved performance and efficiency. We found this setting to work more efficiently than using the images\rq{} original resolution since it would have drastically reduced the field of view for the FCN when processing each sub-volume. In this work, we utilized 3D U-Net for the segmentation of abdominal CTs. However, the proposed hierarchical approach should in principle also work well for other 3D CNN/FCN architectures and 3D image modalities. 
%%%%%%%%%%%%%%%%%%%%%%%%%%%%%%%%%%%%%%%%%%%%%%%%%%%%%%%%%%%%%

In conclusion, we showed that a hierarchical deployment of 3D CNN based on a fully convolutional architecture (3D U-Net) can produce competitive results for multi-organ segmentation on a clinical CT dataset while being efficiently deployed on a single GPU. An overlapping tiles approach during testing produces better results with only moderate additional computational cost. Our method compares favorably to recent state-of-the-art work on a completely unseen dataset.
%%%%%%%%%%%%%%%%%%%%%%%%%%%%%%%%%%%%%%%%%%%%%%%%%%%%%%%%%%%%%
%%%%%%%%%%%%%%%%%%%%%%%%%%%%%%%%%%%%%%%%%%%%%%%%%%%%%%%%%%%%%
\paragraph{\textbf{Acknowledgments}} This work was supported by MEXT KAKENHI (26108006 and 15H01116).
%%%%%%%%%%%%%%%%%%%%%%%%%%%%%%%%%%%%%%%%%%%%%%%%
% BIBLIOGRAPHY
%%%%%%%%%%%%%%%%%%%%%%%%%%%%%%%%%%%%%%%%%%%%%%%%
%\small
\bibliographystyle{chicago} 
\bibliography{references}
\end{document}